\documentclass{tlp}
\usepackage{aopmath}

\newtheorem{definition}{Definition} 
\newtheorem{example}{Example} 
\newtheorem{remark}{Remark} 

\usepackage{epsfig}
\usepackage{url}
\usepackage{amsfonts}
\usepackage{latexsym}

\newcommand{\CASP}{${ASP}^A$}

\newcommand{\mydot}{\textit{.}}

\def\qed{\hfill{$\Box$}}
\def\naf{ not \;}

\submitted{17 June 2005 }
\revised{27 October 2005}
\accepted{13 January 2006}

\title[A Constructive Semantic Characterization of Aggregates
	in ASP]
	{A Constructive Semantic Characterization of Aggregates
	in Answer Set Programming}

\author[Tran Cao Son and Enrico Pontelli]
{TRAN CAO SON and ENRICO PONTELLI\\
	Department of Computer Science\\
	New Mexico State University\\
	\{tson,epontell\}@cs.nmsu.edu}

\begin{document}

\maketitle

\begin{abstract}
This technical note describes a monotone and
continuous fixpoint operator
to compute the answer sets of programs with aggregates.
The fixpoint operator
relies on  the notion of
{\em aggregate solution}.
Under certain conditions, this operator behaves identically
to the three-valued immediate consequence operator
$\Phi^{aggr}_P$ for aggregate programs, independently
proposed in~\cite{Pelov04,PelovDB04}.
This operator  allows us to closely tie the 
computational complexity of the answer set checking
and answer sets existence problems
to the cost of checking a solution of the aggregates
in the program. Finally, we relate the semantics described 
by the operator to other  proposals for 
logic programming with aggregates.
\end{abstract}

\section{Introduction}
Several semantic characterizations of answer sets of logic 
programs with aggregates have been proposed 
over the years (e.g., 
\cite{KempS91,MumickPR90,a-prolog,faberLP04,PelovDB04}).
Most of these proposals have their roots in the 
answer set semantics of normal logic programs 
without aggregates \cite{GL88}. Nevertheless, 
it is known that a straightforward 
generalization of the definition of  answer sets to programs with 
aggregates may  yield {\em non-minimal} and/or
unintuitive answer sets. Consider  the 
following example.  

\begin{example} \label{exp2}
Let $P$ be the program 
\[
\begin{array}{lcl}
  p(1) \leftarrow  & \hspace{1cm} &
  p(2) \leftarrow  \hspace{4cm}
  p(3) \leftarrow  \\
  p(5)  \leftarrow  q &&
   q  \leftarrow   \textnormal{\sc Sum}(\{X \mid p(X)\}) > 10
\end{array}
\]
The aggregate $ \textnormal{\sc Sum}(\{X \mid p(X)\}) > 10$
is satisfied by  any interpretation $M$ of $P$ 
where the sum of $X$ such that 
$p(X)$ is true in $M$ is greater than 10.

A straightforward extension of the original definition of 
answer sets~\cite{GL88}  defines $M$ to be  
an answer set of $P$ if and only if $M$ is the minimal 
model of the reduct $P^M$, where $P^M$ is the program 
obtained by {\em (i)} removing from $P$ all the rules containing in their
body at least an aggregate or a negation-as-failure literal which 
is false in $M$; and {\em (ii)} removing all the aggregates 
and negation-as-failure literals from the remaining rules. 
Effectively, this definition treats aggregates in the same
fashion as negation-as-failure literals.

It is easy to see that for 
$A  = \{p(1),p(2),p(3)\}$ and $B = \{p(1),p(2),p(3),p(5),q\}$, \[
\begin{array}{lll}
P^A = \left\{
\begin{array}{l}
  p(1) \leftarrow  \\
  p(2) \leftarrow  \\
  p(3) \leftarrow \\
     p(5)  \leftarrow   q  \\
\end{array}
\right \}
& \hspace*{0.5cm} & 
P^B = \left \{
\begin{array}{l}
  p(1) \leftarrow \\
  p(2) \leftarrow \\
  p(3) \leftarrow \\
     p(5)  \leftarrow   q \\
   q \leftarrow \\
\end{array}
\right \}
\end{array}
\]
and $A$ and $B$ are minimal model of $P^A$
and $P^B$ respectively. Thus, both $A$ and $B$ are 
answer sets of $P$. As we can see,  treating 
aggregates like negation-as-failure literals  
yields non-minimal answer sets. Accepting $B$ as an
answer set seems counter-intuitive, since $p(5)$ ``supports''
itself through the aggregate.
\qed
\end{example}
Different approaches have been proposed to deal with this problem. Early works
concentrate on finding syntactic (e.g., stratification 
\cite{MumickPR90,KempS91}) and semantic (e.g., monotonic aggregates 
\cite{RossS97,KempS91})
restrictions on aggregates 
which guarantee minimality, and often uniqueness, of answer sets.

In this technical  note, we present a fixpoint operator that allows
us to compute answer sets of normal logic programs with 
{\em arbitrary aggregates}. 
It is a straightforward extension of the Gelfond-Lifschitz definition,
making use of the \emph{same} notion of reduct as in \cite{GL88}, and 
relying on a continuous fixpoint operator for computing 
selected minimal models of the reduct (corresponding to 
our notion of answer sets). This fixpoint operator is a 
natural extension of the traditional
immediate consequence operator $T_P$ to programs with aggregates.
It takes into consideration 
the provisional answer set while trying to verify that it is 
an answer set. This fixpoint operator makes use of
the notion of \emph{aggregate solutions}, and it
captures the 
\emph{unfolding semantics} for normal logic programs with 
aggregates, originally proposed in~\cite{ElkabaniPS04} and completely
developed in~\cite{SonPE05}. This semantics builds on the principle
of unfolding of intensional set constructions, as developed 
in~\cite{DovierPR01}.
This operator corresponds to the $\Phi_P^{aggr}$ operator proposed
in~\cite{PelovDB04,Pelov04}, when ultimate approximating 
aggregates are employed and 2-valued stable models are considered. 
In particular, the two operators are identical when they are applied
to the construction of a correct answer set $M$.

The proposed fixpoint operator allows us also to easily 
demonstrate the existence of a 
large class of logic programs with aggregates 
(which includes recursively defined aggregates 
and non-monotone aggregates)
 for which the problems of answer set checking and of
determining the existence of an answer set is in {\bf P} and 
{\bf NP} respectively. 
Finally, we relate our work to recently proposed
semantics for programs with aggregates
\cite{faberLP04,PelovDB04,SonPE05}.

\section{Preliminary Definitions}
\label{sec2} 

\subsection{Language Syntax}
Let us consider a signature 
$\Sigma_L = \langle {\cal F}_L\cup {\cal F}_{Agg}, {\cal V}\cup {\cal V}_l, \Pi_L \rangle$, 
where ${\cal F}_L$ is a 
collection of constants, ${\cal F}_{Agg}$ is a collection of
unary function symbols, 
${\cal V} \cup {\cal V}_l$ is a denumerable collection of variables
(such that ${\cal V} \cap {\cal V}_l = \emptyset$), 
and $\Pi_L$ is a collection of predicate symbols. In the rest of this paper, we 
will always assume that the set $\mathbb{Z}$ of the integers is
a subset of ${\cal F}_L$---i.e., there are distinct constants representing the 
integer numbers. 
We will refer to  $\Sigma_L$ as the \emph{ASP signature}. 
We will also refer to 
$\Sigma_P = \langle {\cal F}_P,{\cal V}\cup {\cal V}_l, \Pi_P\rangle$
as the \emph{program signature},
where ${\cal F}_P \subseteq {\cal F}_L$,  $\Pi_P \subseteq \Pi_L$, and
${\cal F}_P$ is finite.
We will denote 
with ${\cal H}_P$ the $\Sigma_P$-Herbrand universe, containing
the ground terms built using symbols 
of ${\cal F}_P$, and
with ${\cal B}_P$ the corresponding $\Sigma_P$-Herbrand base. An
ASP-atom is an atom of the form 
$p(t_1,\dots,t_n)$, 
where $t_i \in {\cal F}_P \cup {\cal V}$ 
and $p\in \Pi_P$; an ASP-literal is either an ASP-atom 
or the negation as failure ($not\:A$) of an ASP-atom. 
We will use the traditional notation $\{t_1,\dots,t_k\}$ to denote an
extensional set of terms, and the notation 
$\{\!\!\{t_1,\dots,t_k\}\!\!\}$ to denote an
extensional multiset (or bag) of terms.

\begin{definition}[Intensional Sets and Multisets]
\label{intensionalset}
An \emph{intensional set} is a set of the form 
$ \{X \:\mid\: p(X_1,\dots,X_k)\} $
where $X\in {\cal V}_l$, $X_i$'s are variables or constants
(in ${\cal F}_P$),
$\{X_1,\dots,X_k\} \cap {\cal V}_l = \{X\}$, 
and $p$ is a $k$-ary predicate in $\Pi_P$.
Similarly, an \emph{intensional multiset} is a multiset
of the form 
$$
\{\!\!\{ X \:\mid\:  \exists Z_1,\dots,Z_r \mydot\: 
p(Y_1,\dots,Y_m)\}\!\!\}
$$
where $\{X,Z_1,\dots,Z_r\}\subseteq {\cal V}_l$, $Y_i$ are variables
or constants (of ${\cal F}_P$), 
$\{Y_1,\ldots,Y_m\} \cap {\cal V}_l= \{X,Z_1,\dots,Z_r\}$,
and $X \notin \{Z_1,\dots,Z_r\}$. 
We call $X$ the \emph{grouped variable},
$Z_1,\dots,Z_r$ the \emph{local variables},  
and $p$ the \emph{grouped predicate} of the intensional set/multiset. 
\end{definition}
Intuitively, in an intensional multiset, we collect
the values of $X$ for which  $p(Y_1,\ldots,Y_m)$ is true, under the 
assumptions that the variables $Z_1,\dots,Z_r$ are locally,
 existentially quantified. Multiple
occurrences of the same value of $X$ can appear. For example, if 
$p(X,Z)$ is true for $X=1, Z=2$ and $X=1, Z=3$, then the multiset
$\{\!\{X\mid\exists Z. p(X,Z)\}\!\}$ will correspond to $\{\!\!\{1,1\}\!\!\}$.
Definition \ref{intensionalset} can be easily extended 
to allow more complex types of sets, e.g., sets with
a tuple as the grouped variable and sets with conjunctions of atoms
as property of the intensional construction.

Observe that the variables from ${\cal V}_l$ are used exclusively
as grouped or local variables in defining intensional sets/multisets, and
they cannot occur anywhere else.

We write $\bar{X}$ to denote $X_1,\ldots,X_n$.

\begin{definition}[Aggregate Terms/Atoms]
\begin{itemize}
\item 
An \emph{aggregate term} is of the form $aggr(s)$, where $s$ is 
an intensional set/multiset, and $aggr \in {\cal F}_{Agg}$ (called the
\emph{aggregate function}). 

\item 
An \emph{aggregate atom} has the form 
$aggr(s) \:\: \texttt{op} \:\: Result$,
where {\tt op}  is a relational 
operator in the set $\{=, \neq, <, >, \leq, \geq\}$ and 
$Result \in {\cal V} \cup (\mathbb{Z} \cap {\cal F}_P)$---i.e.,
it is either a variable or a numeric constant.
\end{itemize}
\end{definition}
In our examples, we will focus on the traditional aggregate functions, e.g.,
{\sc Count, Sum, Min}. 
For an aggregate atom $\ell$ of the form $aggr(s) \:\: \texttt{op} \:\: Result$, 
we refer to the grouped variable and predicate of $s$  
as the grouped variable and predicate of $\ell$. 
The set of ASP-atoms constructed from the grouped predicate of $\ell$ and
the terms in ${\cal H}_P$ is denoted by
${\cal H}(\ell)$.

\begin{definition}
[\CASP\ Rule/Program]
\begin{itemize}
\item An \CASP\ rule is of the form
\begin{equation} \label{agg-rule}
A \leftarrow C_1,\ldots,C_m,A_1,\ldots,A_n,\naf B_1, \dots, \naf B_k
\end{equation}
where $A$, $A_1$, $\dots$, $A_n$, $B_1$, $\dots$, $B_k$ are ASP-atoms, while
$C_1, \dots, C_m$ are aggregate atoms ($m\geq 0$, $n\geq 0$, $k\geq 0$). 

\item 
An \CASP\ program is a finite collection of \CASP\ rules.
\end{itemize}
\end{definition}
For an  \CASP\ rule $r$ of the form (\ref{agg-rule}), 
$head(r)$, $agg(r)$, $pos(r)$, and 
$neg(r)$ denote respectively $A$, $\{C_1, \ldots, C_m\}$, 
$\{A_1, \ldots, A_n\}$, and $\{B_1, \ldots, B_k\}$.
Furthermore, $body(r)$ denotes the right-hand side
of the rule $r$.

Observe that  grouped and local  variables in an aggregate atom 
$\ell$ have a scope limited to $\ell$. As such, given an \CASP\ rule, 
it is always possible to rename such variables occurring in
the aggregate atoms $C_1,\ldots,C_m$ apart,  
so that they are pairwise different. Observe also that the
grouped and local variables represent the only occurrences
of variables from ${\cal V}_l$, thus they will not 
occur in $A$, $A_1$, $\dots$, $A_n$, $B_1$, $\ldots$, $B_k$. 
For this reason, without loss of generality, 
whenever we refer to an \CASP\ rule $r$, we will assume 
that the grouped and local variables of its aggregate atoms are 
pairwise different and do not appear in the rest of the rule.

Given a term, literal, aggregate atom, rule $\alpha$, let 
us denote with $fvars(\alpha)$ the set
of variables from ${\cal V}$ present in $\alpha$. The entity
$\alpha$ is ground
if $fvars(\alpha) = \emptyset$.

A ground substitution $\sigma$ is a set $\{X_1/c_1,\ldots,X_n/c_n\}$
where $X_i$'s are distinct variables from ${\cal V}$ 
and $c_i$'s are constants in ${\cal F}_P$. 
For an ASP-atom $p$ (an aggregate atom $\ell$), 
$p \sigma$ ($\ell \sigma$) denotes the ASP-atom 
(the aggregate atom) which is obtained from 
$p$ ($\ell$) by simultaneously replacing 
every occurrence of $X_i$ with $c_i$. 

Let  $r$ be a rule of the form (\ref{agg-rule}) and 
$\{X_1,\ldots,X_t\}$ be the set of free variables occurring in 
$A$, $C_1,\ldots,C_m$,  
$A_1,\ldots,A_n$, and $B_1,\ldots,B_k$---i.e., 
$fvars(r) = \{X_1,\dots,X_t\}$.    
Let $\sigma$ be a ground substitution $\{X_1/c_1,\ldots,X_t/c_t\}$. 
The ground instance of $r$ w.r.t. $\sigma$,
denoted by $r\sigma$,
is the ground rule obtained from $r$ by simultaneously replacing 
every occurrence of $X_i$ with $c_i$. 

By $ground(r)$ 
we denote the set of all ground instances of the rule $r$. For a program $P$, 
the set of all ground instances 
of the rules in $P$, denoted by $ground(P)$, is called the ground 
instance of $P$, i.e., 
$ground(P) = \bigcup_{r \in P} ground(r)$.

\subsection{Aggregate Solutions}

In this subsection we provide the basic definitions of satisfaction and
solution of an aggregate atom. 
\begin{definition}[Interpretation Domain and Interpretation]
The domain of our interpretations is the set 
${\cal D} = {\cal H}_P \cup 2^{{\cal H}_P} \cup {\cal M}({\cal H}_P)$, where
$2^{{\cal H}_P}$ is the set of (finite) subsets of ${\cal H}_P$ and
${\cal M}({\cal H}_P)$ is the set of finite multisets of elements
from ${\cal H}_P$.
An interpretation $I$ is a pair $\langle {\cal D}, (\cdot)^I\rangle$, where
$(\cdot)^I$ is a function that maps ground terms to elements of $\cal D$ and ground
atoms to truth values.
\end{definition}

\begin{definition}[Interpretation Function]
Given a constant $c$, its interpretation $c^I$ is equal to $c$.

\smallskip
\noindent
Given a ground intensional set $s$ of the form $\{X \mid p(\bar{X})\}$,
its interpretation $s^I$ 
is the  set $\{a_1,\dots,a_n\} \subseteq {\cal H}_P$, where 
$(p(\bar{X}))\{X / a_i\}^I$ is equal to true for 
$1\leq i\leq n$, and no other value for $X$ has such property.

\smallskip
\noindent
Given a ground intensional multiset  $s$ of the form 
$\{\!\!\{X\:\mid\: \exists \bar{Z}\mydot p(\bar{X},\bar{Z})\}\!\!\}$,
its interpretation $s^I$ is the multiset
 $\{\!\!\{a_1,\dots,a_k\}\!\!\} \in {\cal M}({\cal H}_P)$ 
where, for each $1\leq i\leq k$, 
there is a ground substitution  $\eta_i$ for $\bar{Z}$ such that 
$p(\bar{X},\bar{Z})\gamma_i^I$ is true for 
$\gamma_i = \eta_i \cup \{X/a_i\}$, and no other elements 
satisfy this property.

\smallskip
\noindent
Given the aggregate term $aggr(s)$, its interpretation is
$aggr^I(s^I)$, where
$$aggr^I: 2^{ {\cal H}_P }\cup {\cal M}({ {\cal H}_P }) \rightarrow \mathbb{Z}.$$

\smallskip
\noindent
Given a ground \CASP\ atom $p(t_1,\dots,t_n)$, its interpretation is
$p^I(t_1^I,\dots,t_n^I)$, where $p^I: {\cal D}^n \rightarrow \{\texttt{true},\texttt{false}\}$.

\smallskip
\noindent
Given a ground aggregate atom  
$\ell$ of the form $aggr(s) \; {\tt op} \; Result$, its
interpretation
$\ell^I$ is true if
$op^I(aggr(s)^I,Result^I)$ is true, where
${\tt op}^I: \mathbb{Z} \times \mathbb{Z} \rightarrow \{\texttt{true},\texttt{false}\}$.
\end{definition}
We will assume that the traditional aggregate functions 
are interpreted in the usual way. E.g.,  {\sc Sum}$^I$ 
is the function that 
maps a set/multiset of numbers to its sum, and 
{\sc Count}$^I$  is the function that maps a set/multiset
of constants to its cardinality. Similarly, we assume that
the traditional relational operators (e.g., $\leq$, $\neq$) are
interpreted according to their traditional meaning. 

Given a literal $not\:p$, its interpretation 
$(not\:p)^I$ is true (false) iff $p^I$ is false (true).

Given an atom, literal, or aggregate atom $\ell$,
we will denote with $I\models \ell$ the fact that 
$\ell^I$ is true.

\begin{definition}[Rule Satisfaction]
$I$ {\em satisfies the body of a ground rule} $r$
(denoted by $I \models body(r)$), if
\begin{list}{}{\topsep=0pt \itemsep=1pt \parsep=0pt}
\item[(i)] $pos(r) \subseteq I$;
\item[(ii)] $neg(r) \cap I = \emptyset$;
\item[(iii)] $I \models c$ for every $c \in agg(r)$.
\end{list}
$I$ {\em satisfies a ground rule} $r$ if $I \models head(r)$ or 
$I \not\models body(r)$.
\end{definition} 
Having specified when an interpretation satisfies an aggregate atom
or a \CASP\ rule, we can define the notion of model of a program.
\begin{definition}[Model]
Let $P$ be an \CASP\ program. An interpretation $M$ is
a {\em model} of $P$ if $M$ satisfies every rule in $ground(P)$.
\end{definition}
In our view of interpretations, we assume that the interpretation
of the aggregate functions and relational operators is fixed. In this
perspective, we will still be able to keep the traditional view of
interpretations as subsets of ${\cal B}_P$. 

\begin{definition}
$M$ is a \emph{minimal model} of $P$ if $M$ is a model of $P$ and 
there is no proper subset of $M$ which is also a model of $P$.
\end{definition}

We will now define a notion called {\em aggregate solution}. 
Observe that the satisfaction of 
an ASP-atom $a$ is \emph{monotonic}, in the sense that if
$I \models a$ and $I \subseteq I'$ then we have that $I' \models a$. 
On the other hand, the satisfaction of an aggregate atom is possibly 
non-monotonic, i.e., $I \models \ell$ 
and $I \subseteq I'$ do not necessarily imply $I' \models \ell$. 
For example, $\{p(1)\} \models \textnormal{\sc Sum}(\{X \mid p(X)\}) \ne 0$ 
but $\{p(1),p(-1)\} \not\models 
\textnormal{\sc Sum}(\{X \mid p(X)\}) \ne 0$. 
The notion of aggregate solution allows us to 
define an operator where
the monotonicity of satisfaction of aggregate 
atoms is used in verifying an answer set. 
\begin{definition}[Aggregate Solution]
Let $\ell$ be a ground aggregate atom. 
An {\em aggregate solution} of $\ell$ is a pair $\langle S_1, S_2 \rangle$ of  
disjoint subsets of ${\cal H}(\ell)$ such that, for every interpretation $I$, 
if $S_1 \subseteq I$ and 
$S_2 \cap I  = \emptyset$ then 
$I \models \ell$. 
${\cal SOLN}(\ell)$ is the set of all the solutions
of  $\ell$.
\end{definition}
It is obvious that 
if $I \models \ell$ then $\langle I \cap {\cal H}(\ell),
				  {\cal H}(\ell) \setminus I \rangle$
is a solution of $\ell$.
Let $S = \langle S_1, S_2 \rangle$ be an aggregate 
solution of an aggregate atom; 
we denote with $S\mydot p$ and $S\mydot n$ the two components 
$S_1$ and $S_2$ of the solution.

\begin{example} \label{ex-sol}
Consider the aggregate atom 
${\textnormal{\sc Sum}(\{X \mid p(X)\}) >  10}$ from the program 
in Example \ref{exp2}. This atom has a unique solution:
$\langle \{p(1), p(2), p(3), p(5)\}, \emptyset\rangle$. 
On the other hand, the aggregate atom 
${\textnormal{\sc Sum}(\{X \mid p(X)\}) >  6}$ has
the following solutions: 
\[
\begin{array}{lrclr} 
\langle \{p(3), p(5)\}, &\emptyset\rangle & \hspace{.8cm} & 
\langle \{p(3), p(5)\}, &\{p(1),p(2)\}\rangle \\
\langle \{p(3), p(5)\}, &\{p(1)\}\rangle & &
\langle \{p(3), p(5)\}, &\{p(2)\}\rangle \\
\langle \{p(2), p(5)\}, &\emptyset\rangle & &
\langle \{p(2), p(5)\}, &\{p(1),p(3)\}\rangle \\
\langle \{p(2), p(5)\}, &\{p(1)\}\rangle & &
\langle \{p(2), p(5)\}, &\{p(3)\}\rangle \\
\langle \{p(1), p(2), p(5)\}, &\emptyset\rangle & &
\langle \{p(1), p(2), p(5)\}, &\{p(3)\}\rangle \\
\langle \{p(1), p(3), p(5)\}, &\emptyset\rangle & &
\langle \{p(1), p(3), p(5)\}, &\{p(2)\}\rangle \\ 
\langle \{p(1), p(2), p(3), p(5)\}, &\emptyset\rangle & &  
\langle \{p(2), p(3), p(5)\}, &\emptyset\rangle  \\
\langle \{p(2), p(3), p(5)\}, &\{p(1)\}\rangle \\
\end{array}
\]
\qed\end{example}
%


\section{A Fixpoint Operator based on Aggregate Solutions}
\label{semantics}

In this section, we construct the semantics for 
\CASP\ programs, through the use of a monotone and
continuous fixpoint operator.
For the sake of simplicity,
 we will assume that programs, ASP-atoms, 
and aggregate atoms referred to in this section 
are ground\footnote{
   A program $P$ with variables can be viewed as a shorthand
   for $ground(P)$.  
}. 
As we will show in Section~\ref{pelov}, this fixpoint operator
behaves as the 3-valued immediate consequence operator of
\cite{PelovDB04} under certain conditions (e.g., use of
ultimate approximating aggregates).

\begin{definition}[Reduct for \CASP\ Programs]
Let $P$ be an \CASP\ program and let $M$ be an interpretation.
The reduct of $P$ with respect to $M$, denoted by $^M\!P$, is defined as
\[
^M\!P=\left\{ head(r) \leftarrow pos(r),agg(r) \mid 
		r \in ground(P), \;\; 
		M \cap  neg(r) = \emptyset 
		\right\}
\]
\end{definition}
Observe that, for a program $P$ without aggregates, the process
of checking whether $M$ is an answer set \cite{GL88} 
requires first computing
the Gelfond-Lifschitz reduct of $P$ w.r.t. $M$ ($P^M$), and then
verifying that $M$ is the least model of $P^M$.
This second step is performed by using 
the van Emden-Kowalski operator $T_{P^M}$ to regenerate $M$, 
by computing the least fixpoint of $T_{P^M}$. I.e.,
we compute the sequence $M_0, M_1, M_2, \dots$ where
$M_0 = \emptyset$ and $M_{i+1} = T_{P_M}(M_i)$.
In every step of regenerating $M$, an atom $a$
is added to $M_{i+1}$ iff there is a rule in $P^M$ whose head is 
$a$ and whose body is contained in $M_i$. 
This process is monotonic, in 
the sense that, if $a$ is added to $M_i$, then $a$ will belong to $M_j$ for 
all $j \ge i$. 

Our intention is to define a $T_P$-like operator for programs with aggregates. 
Specifically, we would like to verify that 
$M$ is an answer set of $P$ by generating a monotone
sequence of interpretations
$M_0 \subseteq M_1 \subseteq \ldots \subseteq M_n \subseteq \ldots = M$.
To do so, we need to specify when a rule of $^M\!P$ can be used, 
i.e., when an ASP/aggregate atom is considered satisfied by $M_i$. 
We also need to ensure that, at each step $i+1$, 
$M_{i+1}$ will {\em still} satisfy all ASP-atoms and 
the aggregate atoms that are satisfied by $M_{i}$. 
 
This observation leads us to define the notion of \emph{conditional 
satisfaction} of an atom (ASP-atom or aggregate atom) over 
a pair of sets of atoms $(I,M)$---where $I$ is an interpretation
generated at some step of the verification process, 
and $M$ is the answer set that needs to be verified.

\begin{definition}[Conditional Satisfaction] \label{cond-sat}
Let $\ell$ be an ASP-atom or an aggregate atom, 
   and $I$, $M$ be two interpretations\footnote{
   Recall that an interpretation is a set of atoms in ${\cal B}_P$.
}. We define the
\emph{conditional satisfaction} of $\ell$ w.r.t. $I$ and $M$, denoted by 
$(I,M) \models \ell$, as:
\begin{list}{$\bullet$}{\topsep=1pt \parsep=0pt \itemsep=1pt}
\item if $\ell$ is ASP-atom, then 
$(I,M) \models \ell \:\:\Leftrightarrow\:\:I\models \ell$

\item if $\ell$ is an aggregate atom, then 
\[ (I,M) \models l \:\:\Leftrightarrow\:\:
	\langle I\cap M\cap {\cal H}(\ell),\: {\cal H}(\ell)\setminus M\rangle
	\textit{ is a solution of $\ell$}\] 	
\end{list}
\end{definition}
The first bullet says that an ASP-atom is satisfied by a pair $(I,M)$ if it is 
satisfied by $I$. The second bullet states that 
$I$ contains enough information of $M$ to guarantee that any successive 
expansion of $I$ towards $M$ will satisfy the aggregate.  
Conditional satisfaction is naturally 
extended to conjunctions of atoms.
The following lemma trivially holds.
\begin{lemma} \label{l0}
Let $\ell$ be an ASP-atom or an aggregate atom 
and $I, J, M$ be interpretations such that
$I \subseteq J$. Then,  $(I,M) \models \ell$ implies $(J,M) \models \ell$. 
\end{lemma}
We are now ready to define the consequence operator for \CASP\
programs.

\begin{definition}[Consequence Operator]
Let $P$ be an \CASP\ program 
and $M$ be an interpretation. 
We define the consequence operator on $P$ and $M$, called 
$K_M^P$, as
\[
K_M^P(I) = \{\:head(r) \:\:\mid\:\: r\in {^M\!P}\:\wedge\: (I,M) \models body(r)\:\}
\]
for every interpretation $I$ of $P$.
\end{definition}
By definition, we have that $K_M^P(I) = T_P(I)$ for definite
programs without aggregate atoms. Thus, $K_M^P$ can be viewed 
as an extension of 
$T_P$ to the class of programs with aggregates. 
The following lemma is a consequence of Lemma~\ref{l0}.
\begin{lemma} \label{l4}
Let $P$ be a program 
and $M$ be an interpretation. Then,
$K_M^P$ is monotone and continuous over the lattice 
$\langle 2^{{\cal B}_P}, \subseteq \rangle$.
\end{lemma}
The above lemma allows us to conclude that the least fixpoint of 
$K_M^P$, denoted by $lfp(K_M^P)$, exists and it is equal to 
$K_M^P \uparrow \omega$. Here, 
$K^P_M \uparrow n$ denotes 
$$\underbrace{K^P_M(K^P_M(\dots(K^P_M}_{n-times \; K^P_M}(\emptyset)\dots)))$$
and $K^P_M \uparrow \omega $ denotes 
$\lim_{n \rightarrow \infty} K^P_M \uparrow n$. 
We are now ready 
to define the concept of \emph{answer set} of an \CASP\ program.

\begin{definition}[Fixpoint Answer Set]
Let $P$ be an \CASP\ program and let $M$ be
an interpretation. $M$ is a \emph{fixpoint answer
set} of $P$ iff $M=lfp(K_M^P)$.
\end{definition}
Whenever it is clear from the context, we will simply talk about
\emph{answer sets} of $P$ instead of fixpoint answer sets.

\begin{example} \label{exp4}
Let us continue with the program $P$ from Example \ref{exp2}.
Since $P$ does not contain negation-as-failure literals, $^M\!P = P$ 
for any interpretation $M$ of $P$.
Any answer set of $P$ must contain $p(1)$, $p(2)$, and 
$p(3)$. We will now show that $A = \{p(1), p(2), p(3)\}$ is the 
unique fixpoint answer set of $P$. It is easy to see that 
\[
\begin{array}{lcl}
K_A^P \uparrow 0  & = & \emptyset\\
K_A^P \uparrow 1 & = & K_A^P (K_A^P \uparrow 0) = \{p(1), p(2), p(3)\} \\
K_A^P \uparrow 2 & = &  \{p(1), p(2), p(3)\} = K_A^P \uparrow 1
\end{array}
\]
Thus, $A$ is indeed a fixpoint answer set of $P$.

Let us consider $B = \{p(1), p(2), p(3),p(5), q\}$. 
We have that $^B\!P = P$ and it is easy to verify that 
$lfp(K_B^P) = \{p(1), p(2), p(3)\}$.
Therefore, $B$ is not a fixpoint answer set of $P$. It is easy to check 
that no proper superset of $A$ is a fixpoint answer set of $P$, i.e., $A$ is 
the unique answer set of $P$.
\qed
\end{example}
In the next example, we show how this definition works when the
programs contain negation-as-failure literals.
\begin{example}
Let $P$ be the program\footnote{
   We would like to thank Vladimir Lifschitz for providing us this example.
}:
\[
\begin{array}{lll}
   p(a) & \leftarrow & \textnormal{\sc Count}(\{X \mid p(X)\}) > 0 \\
   p(b) & \leftarrow & \naf q \\
   q    & \leftarrow & \naf p(b) \\
\end{array}
\]
We will show now that the program has two answer sets $A = \{q\}$
and $B = \{p(b),p(a)\}$. We have that 
\begin{itemize}
\item $^A\!P$ consists of the first rule and the fact 
$q$. The verification that $A$ is an answer set of $P$ is shown next. 
\[
\begin{array}{lcl}
K_A^P \uparrow 0 & = & \emptyset\\
K_A^P \uparrow 1  & = & K_A^P (K_A^P \uparrow 0) = \{q\}\\
K_A^P \uparrow 2 & = & \{q\} = K_A^P\uparrow 1
\end{array}
\]
$p(a)$ cannot belong to $K_A^P \uparrow 1$ since 
$\langle \emptyset, \emptyset \rangle$ is not a solution of the 
aggregate atom $\textnormal{\sc Count}(\{X \mid p(X)\}) > 0$.

\item $^B\!P$ consists of the first rule and the fact 
$p(b)$.   
\[
\begin{array}{lcl}
K_B^P \uparrow 0 & = & \emptyset\\
K_B^P \uparrow 1  & = &  K_B^P (K_B^P \uparrow 0) = \{p(b)\}\\
K_B^P \uparrow 2 & = & \{p(b),p(a)\}  \\
K_B^P \uparrow 3 & = & \{p(b),p(a)\} = K_B^P \uparrow 2 
\end{array}
\]
$p(a)$ belongs to $K_B^P \uparrow 2$  since
$\langle \{p(b)\}, \emptyset \rangle$ is a solution of the 
aggregate atom $\textnormal{\sc Count}(\{X \mid p(X)\}) > 0$.
\end{itemize}
It is easy to see that $P$ does not have any other answer sets.
\qed 
\end{example}

\section{Related Work and Discussion}
\label{discuss}

In this section, we will relate our proposal to the unfolding
semantics presented in \cite{SonPE05} and to 
two other recently proposed semantics for 
programs with aggregates\footnote{
  A detailed comparison between the semantics in \cite{SonPE05} 
  and earlier proposals for programs with aggregates 
  can be found in the same report. 
}---i.e., 
the ultimate stable model semantics 
\cite{PelovDB03,PelovDB04,Pelov04} and the minimal answer set semantics 
\cite{faberLP04}. We will also investigate some of the computational 
complexity issues related to 
determining the fixpoint answer sets of \CASP\ programs.


\subsection{Equivalence of Fixpoint Semantics and Unfolding Semantics}
We will show that the notion of fixpoint answer set
corresponds to the {\em unfolding semantics} 
presented in \cite{SonPE05}. 
To make this note self-contained, let us 
recall the basic definition of the unfolding semantics. 
For a ground aggregate atom $c$  and an 
interpretation $M$, let 
$${\cal S}(c, M) = \left\{S_c 
\:\mid\: S_c \in {\cal SOLN}(c),
 \:S_c\mydot p \subseteq M,\: S_c \mydot n \cap M = \emptyset \right\}
$$
Intuitively, ${\cal S}(c, M)$ is the set of solutions of $c$ 
which are satisfied by $M$. 
For a solution $S_c \in {\cal S}(c,M)$,
the unfolding of $c$ in $M$ w.r.t. $S_c$ 
is the conjunction $\bigwedge_{a \in S_c\mydot p} a$. 
We say that $c'$ is an unfolding of $c$
with respect to $M$ if $c'$ is an unfolding of $c$ in $M$ with respect 
to some $S_c \in {\cal S}(c, M)$. When 
${\cal S}(c, M) = \emptyset$, we say that $false$ 
is the unfolding of $c$ in $M$. The unfolding of 
a rule $r \in ground(P)$ with respect to $M$
is the set of rules $unfolding(r,M)$ defined 
as follows:
\begin{enumerate}
\item If  
	$neg(r) \cap M \ne \emptyset$  or
	there is some $c \in agg(r)$ 
		such that $false$ is the unfolding of $c$ 
		in $M$
	then $unfolding(r,M) = \emptyset$;
\item If $neg(r) \cap M = \emptyset$ and 
	$false$ is not the unfolding of $c$ 
	for every $c \in agg(r)$,
	then $r' \in unfolding(r,M)$ where
	\begin{enumerate}
	\item $head(r') = head(r)$
	\item $neg(r') = neg(r)$
	\item there is a sequence of aggregate solutions 
		$\langle S_c \rangle_{c \in agg(r)}$ 
		for the aggregates in $agg(r)$, 
		such that $S_c \in {\cal S}(c,M)$ for every 
		$c \in agg(r)$ and 
		$pos(r') = pos(r) \cup \bigcup_{c \in agg(r)} S_c\mydot p$.
	\end{enumerate}
\end{enumerate}
For a program $P$, $unfolding(P,M)$ denotes the set
of unfolding rules of $ground(P)$ w.r.t. $M$. 
$M$ is an \CASP-answer set of $P$ iff 
$M$ is an answer set of $unfolding(P, M)$.

This notion of unfolding derives from the work on
unfolding of intensional sets~\cite{DovierPR01}, and has been
independently described in~\cite{PelovDB03}.

\begin{lemma}\label{l1}
Let $c$ be an aggregate atom, let $M$ be an interpretation,
and let $S_c$ be a solution of $c$ such that $S_c \in {\cal S}(c,M)$.
Then, $\langle S_c\mydot p, {\cal H}(c)\setminus M\rangle$ is a solution
	of $c$.
\end{lemma}
\begin{proof}
Let us consider an interpretation $I$ such that 
$S_c\mydot p \subseteq I$ and $I \cap ({\cal H}(c)\setminus M) = \emptyset$. 
Because $S_c\mydot n \subseteq {\cal H}(c) \setminus M$, 
$I \cap S_c\mydot n = \emptyset$. Since $S_c$ is a solution,
$I\models c$. Since this holds for every interpretation $I$
satisfying 
$S_c\mydot p\subseteq I$ and $I \cap ({\cal H}(c)\setminus M) = \emptyset$, 
we have that $\langle S_c\mydot p, {\cal H}(c)\setminus M\rangle$ is a solution
	of $c$. 
\end{proof}

\begin{lemma} \label{l2}
Let $R=unfolding(P,M)$. Then
$T_R\uparrow i = K_M^P\uparrow i$ for $i\geq 0$.
\end{lemma}
\begin{proof}
Let us prove the result by induction on $i$.

\noindent
{\em Base:} for $i=0$, we have that $T_R \uparrow 0  =  \emptyset   =  K_M^P\uparrow 0$,
and the result is obviously true.
Let us consider the case $i=1$. 
\begin{itemize}
\item Let  $p \in T_R\uparrow 1 = \{
			\ell \:\mid\: (\ell\leftarrow) \in R\}$.
	If $p\leftarrow$ is a fact in $P$, then it is also
	a fact in $^M\!P$. This means that $p\leftarrow$ is an
	element of $^M\!P$, and thus $p$ is in
	$K_M^P\uparrow 1$.
	Otherwise, there is a rule $r$ in $P$, such that
	\begin{list}{-}{\topsep=1pt \parsep=0pt \itemsep=1pt}
	\item $head(r) = p$;
	\item $pos(r) = \emptyset$;
	\item $neg(r) \cap M = \emptyset$; and 
	\item for each $\ell \in agg(r)$ we have that 
		there exists a solution of $\ell$ of the
		form $\langle \emptyset,J\rangle$ such that
		$M \cap J =\emptyset$.
	\end{list}
	The rule $p\leftarrow agg(r)$ is a 
	rule in $^M\!P$. From Lemma~\ref{l1} we can conclude
	that $(\emptyset,M) \models agg(r)$, thus ensuring
	that $p \in K^P_M\uparrow 1$.

\item Let $p \in K^P_M\uparrow 1$. Thus, there exists
	a rule $r'\in {^M\!P}$ such that $(\emptyset,M)\models body(r)$
	and $head(r')=p$.
	This means that there is a rule $r \in P$ such that
	\begin{list}{-}{\topsep=1pt \parsep=0pt \itemsep=1pt}
	\item $head(r) = head(r') = p$;
	\item $M \cap neg(r) = \emptyset$;
	\item $pos(r) = \emptyset$; and
	\item $agg(r) = agg(r')$.
	\end{list}
	Since $(\emptyset,M) \models agg(r)$, we have that,
	for each $c\in agg(r)$,
	$\langle \emptyset, {\cal H}(c)\setminus M\rangle$ is
	a solution of $c$. This means that the rule
	$p \leftarrow $ is in $unfolding(P,M)$. This also means
	that $p\in T_R\uparrow 1$.
 \end{itemize}

\smallskip

\noindent
\emph{Step:} Let us assume that the result holds for $i \leq k$ and
	consider the iteration $k+1$.
	
\begin{list}{$\bullet$}{\topsep=1pt \parsep=0pt \itemsep=1pt \leftmargin=10pt}
\item Let $p \in T_R\uparrow (k+1)$ and 
	$p \not\in T_R\uparrow k$.
	Thus,  there is a rule $r'$ in $R$ such 
	that
	\begin{list}{-}{\topsep=1pt \parsep=0pt \itemsep=1pt}
	\item $head(r')=p$; and 
	\item $pos(r') \subseteq T_R\uparrow k$.
	\end{list}
	This implies that there is a rule $r\in P$ such that
	\begin{list}{-}{\topsep=1pt \parsep=0pt \itemsep=1pt}
	\item $head(r) = p$;
	\item $pos(r) \subseteq T_R\uparrow k$;
	\item $M \cap neg(r) =\emptyset$; and 
	\item for each $c \in agg(r)$, there is a solution
		$S_c$ s.t. $S_c\mydot p \subseteq T_R\uparrow k$
		and $M \cap S_c\mydot n = \emptyset$.
	\end{list}
	This also means that 
	$p \leftarrow pos(r),agg(r)$ is a rule in
	$^M\!P$. 

	We already know that $pos(r) \subseteq K^P_M\uparrow k$.
	Now we wish to show that $(K^P_M\uparrow k, M) \models agg(r)$.
	Lemma \ref{l1} shows that, for each $c \in agg(r)$, 
	$\langle S_c\mydot p, {\cal H}(c)\setminus M\rangle$ is a solution of $c$. 
	This allows us to conclude that 	$p \in K^P_M\uparrow (k+1)$.

\item Let  $p \in K^P_M\uparrow (k+1)$ and
	$p \not\in K^P_M\uparrow k$.
		This means that there is a rule $r'$ in $^M\!P$ such that
	\begin{list}{-}{\topsep=1pt \parsep=0pt \itemsep=1pt}
	\item $head(r') = p$;
	\item $pos(r') \subseteq K^P_M\uparrow k$; and 
 	\item $( K^P_M\uparrow k, M) \models body(r')$
	\end{list}
	This also means that there is  a rule $r$ in $P$ such 
	that 
	\begin{list}{-}{\topsep=1pt \parsep=0pt \itemsep=1pt}
	\item $head(r) = head(r') = p$;
	\item $agg(r) = agg(r')$;
	\item $pos(r) = pos(r')$;
	\item $neg(r) \cap M = \emptyset$; and 
	\item for each $c \in agg(r)$, 
		$S_c = \langle K^P_M\uparrow k\cap M\cap {\cal H}(c),
			{\cal H}(c) \setminus M \rangle$ is a 
	solution of $c$. 
	\end{list}
	This means that there is a rule $r''$ in 
	$unfolding(P,M)$ such that:
	\begin{list}{-}{\topsep=1pt \parsep=0pt \itemsep=1pt}
	\item $head(r'') = p$
	\item $pos(r'') = pos(r) \cup \bigcup_{c\in agg{r}} S_c\mydot p$
	\end{list}
	Since each $S_c\mydot p \subseteq K^P_M\uparrow k=T_R\uparrow k$
	for each $c \in agg(r)$
	and $pos(r) \subseteq   K^P_M\uparrow k=T_R\uparrow k$, 
	we have that 
	$p \in T_R\uparrow (k+1)$.
\end{list}
\end{proof}

\begin{theorem} \label{fixpointasp}
Let $P$ be a program with aggregates. $M$ is an answer set
of $unfolding(P,M)$ iff $M$ is a fixpoint answer set of $P$. 
\end{theorem}
\begin{proof}
Let $R = unfolding(P,M)$. We have that 
$M$ is an answer set of $P$ iff
$M = T_R\uparrow \omega$ iff 
$M = K^P_M \uparrow \omega$ (Lemma \ref{l2}). 
\end{proof}

The results from~\cite{SonPE05} and Theorem~\ref{fixpointasp} provide us
a direct connection between fixpoint answer sets and other 
semantics for logic programs with aggregates. 

\subsection{Faber et al.'s Minimal Model Semantics}
The notion of answer set proposed in \cite{faberLP04}
is based on a new notion of reduct, defined as follows. 
Given a program $P$ and a set of ASP-atoms $M$, 
the {\em reduct of P with respect to M}, denoted by 
$\Gamma(M,P)$, is obtained by removing from 
$ground(P)$ those rules whose body cannot be satisfied by $M$.
In other words, $\Gamma(M,P) = \{r \mid r \in ground(P), M \models body(r)\}$.

\begin{definition} [FLP-answer set, \cite{faberLP04}]
\label{d-faberLP04}
For a program $P$, $M$ is an {\em FLP-answer set} of $P$ 
if it is a minimal model of $\Gamma(M,P)$.  
\end{definition}
The following theorem derives directly from Theorem~\ref{fixpointasp} and
\cite{SonPE05}.

\begin{theorem}
\label{th2}
Let  $P$ be a program with aggregates. If $M$ is a fixpoint answer set, 
then $M$ is an FLP-answer set of $P$.
\end{theorem}
Observe that there are cases where FLP-answer sets are not fixpoint answer
sets.

\begin{example}
Consider the program $P$ where 
\[
\begin{array}{lll}
p(1)& \leftarrow & \textnormal{\sc Sum}(\{X \mid p(X)\}) \ge 0 \\
p(-1)& \leftarrow & p(1) \\
p(1)& \leftarrow & p(-1) \\
\end{array}
\]
It can be checked that $M = \{p(1), p(-1)\}$ is an FLP-answer set of $P$. 
It is possible to show that 
$\textnormal{\sc Sum}(\{X \mid p(X)\}) \geq 0$ has the 
following solutions: 
$\langle \emptyset, \{p(1),p(-1)\} \rangle$,
$\langle \{p(1)\}, \{p(-1)\} \rangle$,
$\langle \{p(1)\}, \emptyset \rangle$, and
$\langle \{p(1),p(-1)\}, \emptyset \rangle$. 

We have that $K^P_M(\emptyset) = \emptyset$ since 
$\langle \emptyset, \emptyset \rangle$ is not a solution 
of $\textnormal{\sc Sum}(\{X \mid p(X)\}) \ge 0$. 
This implies that $lfp(K^P_M) = \emptyset$.
Thus, $M$ is not a fixpoint answer set of $P$. 
It can be easily verified that $P$ does not have any fixpoint answer 
set. 
\qed
\end{example}

\begin{remark} 
If we replace in $P$ the rule
$p(1) \leftarrow \textnormal{\sc Sum}(\{X \mid p(X)\}) \ge 0$ 
with the intuitively equivalent {\sc Smodels} weight constraint 
rule $$p(1) \leftarrow 0 [p(1) = 1, p(-1)=-1]$$ 
we obtain a program that does not have answer sets 
in {\sc Smodels}. 
\end{remark}

The above example shows that our characterization
differs from  \cite{faberLP04}. Our definition is closer to 
{\sc Smodels}' understanding of aggregates.

\subsection{Approximation Semantics for Logic Programs with Aggregates}
\label{pelov}

The work of Pelov et al. \cite{PelovDB03,Pelov04,PelovDB04} contains an elegant
generalization of several semantics of logic programs to logic 
programs with aggregates. The key idea in this work is the use of 
approximation theory in defining several semantics for logic programs 
with aggregates (e.g., two-valued semantics, ultimate three-valued stable 
semantics, three-valued stable model semantics). 
In particular, in \cite{PelovDB04}, the authors describe a fixpoint
operator, called $\Phi^{appr}_P$, operating on 3-valued interpretations and
parameterized by the choice of approximating aggregates.

It is possible to show the following results:
\begin{itemize}
\item Whenever the approximating aggregate used in $\Phi^{appr}_P$
	is the \emph{ultimate approximating aggregate}~\cite{PelovDB04},
	then the fixpoint semantics defined by the 
	operator $K^P_M$ coincides with the two-valued stable model semantics
	defined by the operator $\Phi^{appr}_P$.
\item It is possible to prove a stronger result,
	 showing that, if $I \subseteq M$
	then $K_M^P(I) = \Phi^{aggr,1}_P(I,M)$, where $\Phi^{aggr,1}_P(I,M)$
	denotes the first component of $\Phi^{aggr}_P(I,M)$. In other words,
	when ultimate approximating aggregates are employed and $M$ is an
	answer set, then  the fixpoint
	operator of Pelov et al. and
	$K_M^P$ behave identically.
\end{itemize}
We will prove next the first of these two results. The proof of the
second result (kindly contributed by one of the anonymous reviewers)
can be found in Appendix A.
 We will make use of the translation of logic programs 
with aggregates to normal logic programs, denoted by $tr$, described
in \cite{PelovDB03}.  
The translation in \cite{PelovDB03} 
and the unfolding described in the previous subsection
are similar\footnote{It should be noted that our translation
        builds on our previous work on semantics of logic
	programming with sets and aggregates 
	\cite{DovierPR01,DovierPR03,ElkabaniPS04}
	and was independently developed w.r.t. the work in
	\cite{PelovDB03}.}.

For the sake of  completeness, we will review the translation 
of \cite{PelovDB03}, presented using the notation of our paper. 
Given a ground logic program with aggregates $P$, $tr(P)$ denotes the 
ground normal logic program obtained after the translation.
The process begins with the translation of each aggregate atom 
$\ell$ of the form $aggr(s) \:\: \texttt{op} \:\: Result$ 
into a disjunction $tr(\ell) = \bigvee F^{{\cal H}(\ell)}_{(s_1,s_2)}$, where 
 $s_1 \subseteq s_2 \subseteq {\cal H}(\ell)$,  and 
each $F^{{\cal H}(\ell)}_{(s_1,s_2)}$ is a conjunction of the form 
\[
\bigwedge_{l \in s_1} l \wedge \bigwedge_{l \in {\cal H}(\ell) 
		\setminus s_2} \naf l 
\]
The construction of $tr(\ell)$ considers only the pairs $(s_1,s_2)$
that satisfy the  following condition:  
each interpretation $I$ such that  $s_1 \subseteq I$ 
and ${\cal H}(\ell) \setminus s_2 \cap I = \emptyset$ must  satisfy $\ell$.
The translation
$tr(P)$ is then created by replacing rules with disjunction 
in the body by a set of standard rules in a straightforward way. For example, 
the rule 
\[
a \leftarrow (b \vee c), d
\]
is replaced by the two rules 
\[
\begin{array}{lcl} 
a \leftarrow b, d  & \hspace{1cm} &
a \leftarrow c, d\\
\end{array}
\]
From the definitions of $tr(\ell)$ and of aggregate solutions, we have the
following simple lemma:
\begin{lemma} \label{tr1}
For every aggregate atom $\ell$ of the form
$aggr(s) \:\: \texttt{op} \:\: Result$,
$S$ is a solution of $\ell$ if and only if 
$F^{{\cal H}(\ell)}_{(S\mydot p,{\cal H}(\ell) \setminus S\mydot n)}$
is a disjunct in $tr(\ell)$.
\end{lemma}
We next show that fixed point answer sets of $P$ are answer sets of $tr(P)$.

\begin{lemma}  
For a program $P$, $M$ is a fixpoint answer set of $P$ 
iff $M$ is an answer set of $tr(P)$. 
\end{lemma}
\begin{proof}
Let $M$ be an interpretation of $P$ and $R = unfolding(P,M)$. 
We have that $R$ is a positive program.
Furthermore, let $Q$ denote the result of the Gelfond-Lifschitz reduction of 
$tr(P)$ with respect to $M$, i.e., $Q = (tr(P))^M$. 
We will prove by induction on $k$ 
that if $M$ is an answer set of $Q$
then $T_Q \uparrow k = T_R \uparrow k$ for every $k \ge 0$. 
The equation holds trivially for $k=0$. Let us consider 
now the case for $k$, assuming that $T_Q \uparrow l = T_R \uparrow l$
for $0 \le l < k$. 

\begin{enumerate}
\item Consider $p \in T_Q \uparrow k$.
This means that there exists some rule $r' \in Q$ such that 
$head(r') = p$ and $body(r') \subseteq T_Q \uparrow (k-1)$. 
$r' \in Q$ if and only if there exists some 
$r \in P$ such that $r' \in tr(r)$. Together with Lemma \ref{tr1}, 
we can conclude that  there exists a sequence of aggregate solutions
$\langle S_c \rangle_{c \in agg(r)}$ for the aggregate atoms in $body(r)$ 
such that $pos(r') = pos(r) \cup \bigcup_{c \in agg(r)} S_c\mydot p$, and 
$(neg(r) \cup \bigcup_{c \in agg(r)} S_c\mydot n) \cap M = \emptyset$.
This implies that $r' \in R$. Together with the 
inductive hypothesis, we can conclude that $p \in T_R \uparrow k$.

\item Consider $p \in T_R \uparrow k$. This implies that 
there exists some rule $r' \in R$ such that 
$head(r') = p$ and $body(r') \subseteq T_R \uparrow (k-1)$. From 
the definition of $R$, we conclude 
that there exists some rule 
$r \in ground(P)$ and a sequence of aggregate solutions
$\langle S_c \rangle_{c \in agg(r)}$ for the aggregate atoms in $body(r)$ 
such that $pos(r') = pos(r) \cup \bigcup_{c \in agg(r)} S_c\mydot p$, and 
$(neg(r) \cup \bigcup_{c \in agg(r)} S_c\mydot n) \cap M = \emptyset$.
Using Lemma \ref{tr1}, we can conclude that $r' \in Q$. 
Together with the 
inductive hypothesis, we can conclude that $p \in T_Q \uparrow k$.
\end{enumerate}
Similar arguments can be used to show that if $M$ is an answer set of
$R$, $T_Q \uparrow k = T_R \uparrow k$ for every $k \ge 0$, which means 
that $M$ is an answer set of $Q$.
\end{proof}

In \cite{PelovDB03}, it is shown that answer sets of $tr(P)$ coincide 
with the \emph{two-valued partial stable models} of $P$ (defined by the 
operator $\Phi^{aggr}_P$). This, together with the above lemma 
and Theorem \ref{fixpointasp}, allows us to conclude the following 
theorem. 

\begin{theorem}
For a program with aggregates $P$, $M$ is an fixpoint answer set of 
$P$ if and only if it is a fixpoint of the operator $\Phi^{aggr}_P$ 
of \cite{PelovDB04}.
\end{theorem}

\subsection{Complexity Considerations}

We will now discuss the complexity of computing 
fixpoint answer sets. In what follows, we will assume that 
the program $P$ is given and it is a ground program whose language
is finite. By the \emph{size} of a program, we 
mean the number of rules and atoms present in it, as in \cite{faberLP04}. 
Observe that, in order to support the computation of the iterations
of the $K^P_M$ operator, we need the ability to determine
whether a given $\langle I,J \rangle$ is a solution
of an aggregate atom. 
For this reason, we classify programs with aggregates 
by the computational complexity of its aggregates. 
We define a notion, called $C$-{\em decidability}, 
where $C$ denotes a complexity class in 
the complexity hierarchy, as follows.

\begin{definition}
Given an aggregate atom $\ell$ and an interpretation $M$,
we say that $\ell$ is $C$-{\em decidable} if 
its truth value with respect to $M$ can 
be decided by an oracle of the complexity $C$. 
A program $P$ is called {\em $C$-decidable} if the aggregate 
atoms occurring in $P$ are $C$-decidable.
\end{definition}

It is easy to see that aggregate atoms built using 
the standard aggregate functions 
({\sc Sum, Min, Max, Count, Avg}) and 
relations ($=,\ne,\geq,>,\leq,<$) are polynomially 
decidable. 
The solution checking problem is defined as follows.
\begin{definition}
[(SCP) Solution Checking Problem]
{\em\bf Given} an aggregate atom $\ell$, its language
extension ${\cal H}(\ell)$, and a pair of disjoint sets
$I, J \subseteq {\cal H}(\ell)$, 
{\em\bf Determine} whether $\langle I,J \rangle$ is a solution of $\ell$.
\end{definition}
We have the following lemma. 
\begin{lemma} \label{l5}
The SCP is in
{\bf co{-}NP}$^C$ for $C$-decidable aggregate atoms.
\end{lemma}
\begin{proof}
We will show that the complexity of the inverse problem of the
SCP is in {\bf NP}$^C$, i.e., 
determining whether 
$\langle I, J \rangle$ is not a solution of $\ell$
is in {\bf NP}$^C$. 

By definition,
$\langle I, J \rangle$ is not a solution of $\ell$ if there exists
an interpretation $M$ 
such that $I \subseteq M$, $J \cap M = \emptyset$,
and $M \not\models \ell$. To answer this question, 
we can guess an interpretation $M$
and check whether $\ell$ is false in $M$.
If it is, we conclude that 
$\langle I, J \rangle$ is not a solution of $\ell$.
Because $\ell$ is $C$-decidable and there are at most 
$2^{|{\cal H}(\ell) \setminus (I \cup J)|}$ interpretations
that can be used in checking whether 
$\langle I, J \rangle$ is not a solution of $\ell$, 
we conclude that the complexity of the inverse problem is 
in {\bf NP}$^C$.
\end{proof}
We will now address the problem of answer set checking and
determining the existence of answer set. 
\begin{definition}
[(ACP) Answer Set Checking Problem]
{\em\bf Given} an interpretation $M$ of $P$,
{\em\bf Determine} whether $M$ is an answer set of $P$.
\end{definition}
\begin{definition}
[(AEP) Answer Set Existence Problem]
{\em\bf Given} a program $P$,
{\em\bf Determine} whether $P$ has a fixpoint answer set.
\end{definition}
The following theorem follows from Lemma \ref{l5}.
\begin{theorem}
The ACP of $C$-decidable programs is in
{\bf co{-}NP}$^C$.
\end{theorem}
\begin{proof}
The main tasks in checking whether $M$ is an answer set of $P$ are {\em (i)} 
computing $^M\!P$; and {\em (ii)} computing $lfp(K_M^P)$. 
Obviously, $^M\!P$ can be constructed in time linear in the size of $P$, 
since the reduction relies on the 
satisfiability test of a negation-as-failure 
literal $\ell$ w.r.t. $M$.
Computing $lfp(K_M^P)$ requires at most $na$ iterations,
i.e., $lfp(K_M^P) = K_M^P \uparrow na$,  
where $na$ is the number of atoms of $P$, each step 
is in {\bf co{-}NP}$^C$, due to the requirement of solution 
checking. 
\end{proof}
This theorem allows us to conclude the following result.
\begin{corollary}
The AEP for $C$-decidable program 
is in {\bf NP}$^{\textnormal{\bf co{-}NP}^C}$. 
\end{corollary}
So far, we discussed the worst case analysis of answer set 
checking and determining the existing of an answer set based on 
a general assumption about the complexity of computing the aggregate functions
and checking the truth value of aggregate atoms. 
Next we analyze the complexity of these problems 
w.r.t. the class of programs whose aggregate atoms 
are built using standard 
aggregate functions and operators. 

\subsubsection{Complexity of Solution Checking for Standard Aggregates}

We will now focus on the class of programs 
defined in Section \ref{sec2} with standard aggregate
functions ({\sc Sum, Min, Max, Count, Avg}) and relations 
($=$, $\geq$, $>$, $\leq$, $<$, $\ne$). It is easy to see that all
aggregate atoms involving these functions and relations 
are {\bf P}-decidable. Therefore, 
by Lemma \ref{l5}, the SCP for standard aggregates will be at 
most {\bf co{-}NP}. We will now show that 
it is {\bf co{-}NP}-complete. 

\begin{theorem} \label{t10}
The SCP for standard aggregates is 
{\bf co{-}NP}-complete. 
\end{theorem}
\begin{proof}
Membership follows from Lemma \ref{l5}. 
To prove hardness, we will translate  
a well-known {\bf NP}-complete problem,
namely the subset sum  problem \cite{cormen}, 
to the complement of the  solution checking problem. 
An instance $Q$ of the subset sum problem is given 
by a set of non-negative integers $S$ and an
integer $t$, and the question is to determine whether 
there exists any non-empty subset $A$ of
$S$ such that $\sum_{x \in A} x = t$. 

Let ${\cal H}(\ell) = \{p(x) \mid x \in S\}$ for some unary predicate $p$. 
We define an instance of the solution checking problem,
$s(Q)$, by setting $I=\emptyset$, $J=\emptyset$, and
$\ell = \textnormal{\sc Sum}(\{X \mid p(X)\}) \ne t$. It is easy to see 
that $s(Q)$ is equivalent to $Q$ as follows:
if $\langle I,J \rangle$ is a solution of $\ell$ then $Q$ 
does not have an answer;
if $\langle I,J \rangle$ is not a solution to $\ell$ then 
$Q$ has an answer. This proves the desired result.
\end{proof}

The above theorem shows that, in general, the inclusion of 
standard aggregates implies that the answer set checking problem and 
the problem of determining the existing of an answer set are
in {\bf co{-}NP} and {\bf NP}$^{\textnormal{\bf co{-}NP}}$ respectively.
Fortunately, there is a large class of programs with 
standard aggregates for which the complexity of these two 
problems are in {\bf P} and {\bf NP} respectively, as shown 
next.

\begin{lemma} \label{l10}
Let $\ell$ be an aggregate of the form 
$\textnormal{\sc Sum}(\{X \mid p(X)\}) = v$, where $v$ is a constant
in  $\mathbf{R}$. 
Let $I,J \subseteq {\cal H}(\ell)$ such that $I \cap J = \emptyset$.
Then, 
determining whether $\langle I, J \rangle$ is a solution of $\ell$
can be done in time polynomial in the size of ${\cal H}(\ell)$. 
\end{lemma}
\begin{proof}
Let us denote with $\pi$ the function that projects an element $p$
of ${\cal H}(\ell)$ to the value that $p$ assigns to the collected
variable. This value will be denoted by $\pi(p)$. We prove the lemma
by providing a polynomial algorithm for determining whether 
$\langle I, J \rangle$ is a solution of $\ell$. 
\begin{center}
\begin{minipage}[t]{.8\textwidth}
{\begin{tabbing} 
iiii\=iiii\=iiiii\=iiii\=iiii\=iiii\=iiii\=iiii\=iiii\kill
1:\>{\bf function} {\tt Check\_Solution} ($v$, $\langle I, J\rangle$, 
		${\cal H}(\ell)$)\\ 
2:\>\> compute $s = \Sigma_{p \in I} \pi(p)$ \\ 
3:\>\> {\bf if} $s \ne v$ {\bf then return false}\\ 
4:\>\> {\bf if} ${\cal H}(\ell) \setminus (I \cup J) = \emptyset$ 
		{\bf then return true};\\ 
5:\>\> {\bf forall} ($p \in {\cal H}(\ell) \setminus (I \cup J)$)\\ 
6:\>\>\> {\bf if} $\pi(p) \ne 0$ {\bf then return false} \\
7:\>\> {\bf endfor} \\
8:\>\> {\bf return true} \\
\end{tabbing}} 
\end{minipage}
\end{center} 
It is easy to see that the above algorithm returns true (resp.
false) if and only if 
$\langle I, J \rangle$ is (resp. is not) a solution of $\ell$. 
Furthermore, the time complexity of the above algorithm 
is polynomial in the size of ${\cal H}(\ell)$. This proves the lemma. 
\end{proof}

The above lemma shows that the solution checking problem can be
solved in polynomial time for a special type of standard aggregate
atoms. Indeed, this can be proven for all standard aggregates but those
of the form $\textnormal{\sc Sum} \ne v$ and
$\textnormal{\sc Avg} \ne v$. 

\begin{lemma}
Let $\ell$ be the aggregate  $agg(s) \;\mathbf{ op }\; v$ 
where $agg \not\in \{\textnormal{\sc Sum, Avg}\}$ 
or $agg \in  \{\textnormal{\sc Sum, Avg}\}$ and
$\mathbf{op}$ is not `$\ne$'. 
Let $I, J \subseteq {\cal H}(\ell)$, $I \cap J = \emptyset$, and 
$v \in \mathbf{R}$. 
Then, checking if
$\langle I, J \rangle$ is a solution of $\ell$ can be done 
in time polynomial in the size of ${\cal H}(\ell)$. 
\end{lemma}
\begin{proof}
The proof can be done similarly to the proof of Lemma \ref{l10}:
for each type of atom, we develop an algorithm, which 
returns true (resp. false) if $\langle I,J\rangle$ 
is (resp. is not) a solution of $\ell$. 
For brevity, we only discuss the steps which need to be done.
It should be noted that each of these steps can 
be done in polynomial time in the size of ${\cal H}(\ell)$,
which implies the conclusion of the lemma.

\begin{itemize}

\item {\sc Sum:} Let $s = \sum_{p\in I} \pi(p)$. All cases can
be handled in time $O(|{\cal H}(\ell)|)$. Let us consider
the various cases for {\bf op}.

\begin{list}{$\bullet$}{\topsep=1pt \parsep=0pt \itemsep=1pt}
\item The case {\bf op} is '=' has been discussed in Lemma \ref{l10}. 

\item For ${\bf op} \in\{\geq, >\}$,  
let $H_1 = \{p \mid p \in {\cal H}(\ell)
	\setminus (I \cup J),\; \pi(p)<0\}$. 
We have that 
$\langle I,J \rangle$ is a solution of $\ell$ if and only if
$s \; \mathbf{op} \; v$ and $\sum_{p \in H_1} \pi(p) + s \; \mathbf{op} \;  v$.

\item For $\mathbf{op} \in \{\leq,<\}$, let 
$H_1 = \{p \mid p \in {\cal H}(\ell)
	\setminus (I \cup J),\; \pi(p)>0\}$. 
We have that 
$\langle I,J \rangle$ is a solution of $\ell$ if and only if
$s \; \mathbf{op} \;  v$ and 
$\sum_{p \in H_1} \pi(p) + s \; \mathbf{op} \;  v$.
\end{list}

\item{\sc Count:} Let $c = |I|$ and 
$H_1 = {\cal H}(\ell) \setminus (I \cup J)$. All cases can be 
handled in time $O(|{\cal H}(\ell)|)$.

\begin{list}{$\bullet$}{\topsep=1pt \parsep=0pt \itemsep=1pt}
\item If $\mathbf{op} \in \{>,\ge\}$, then  
$\langle I, J\rangle$ is a solution of $\ell$ if and only if 
$c \; \mathbf{op} \; v$. 

\item If  $\mathbf{op} \in \{=,<,\le\}$, then   
$\langle I, J\rangle$ is a solution of $\ell$ if and only if 
$c \; \mathbf{op} \; v$ 
and $c + |H_1| \; \mathbf{op} \; v$.

\item If  $\mathbf{op}$ is $\ne$, then 
$\langle I, J\rangle$ is a solution of $\ell$ if and only if 
either {\em (i)} $|I| > v$;  or 
{\em (ii)} $|I| < v$ and $|H_1| < v-|I|$.
\end{list}

\item{\sc Min:} Let $c = \min  \{\pi(p) \mid p \in I\}$ and 
$c_1 = \min \{\pi(p) \mid p \in {\cal H}(\ell) \setminus (I \cup J)\}$. 
All cases can  be 
handled in time $O(|{\cal H}(\ell)|)$.

\begin{list}{$\bullet$}{\topsep=1pt \parsep=0pt \itemsep=1pt}
\item If $\mathbf{op}$ is $=$ then 
we have that 
$\langle I,J \rangle$ is a solution of $\ell$ if and only if
$c = v $ and $c_1 \ge v$.

\item If  $\mathbf{op} \in \{\leq,<\}$ then  
$\langle I,J \rangle$ is a solution of $\ell$ if and only if
$c \; \mathbf{op} \; v $.

\item If  $\mathbf{op} \in \{\geq,>\}$ then  
$\langle I,J \rangle$ is a solution of $\ell$ if and only if
$c \; \mathbf{op} \; v$ and
$c_1 \; \mathbf{op} \; v$.

\item If $\mathbf{op}$ is $\ne$ then 
$\langle I,J \rangle$ is a solution of $\ell$ if and only if
either {\em (i)} $c < v$;  or 
{\em (ii)} $c > v$ and for every $p \in H_1$, 
$\pi(p) \ne v$.
\end{list}

\item{\sc Max:}
Let $c = \max  \{\pi(p) \mid p \in I\}$ and 
$c_1 = \max \{\pi(p) \mid p \in {\cal H}(\ell) \setminus (I \cup J)\}$.
All cases can  be 
handled in time $O(|{\cal H}(\ell)|)$.

\begin{list}{$\bullet$}{\topsep=1pt \parsep=0pt \itemsep=1pt}
\item If $\mathbf{op}$ is $=$ then 
$\langle I,J \rangle$ is a solution of $\ell$ if and only if
$c = v $ and $c_1 \le v$.

\item If $\mathbf{op} \in \{\geq,>\}$ then
$\langle I,J \rangle$ is a solution of $\ell$ if and only if
$c \; \mathbf{op} \; v $.

\item If $\mathbf{op} \in \{\leq,<\}$ then  
$\langle I,J \rangle$ is a solution of $\ell$ if and only if
$c \; \mathbf{op} \; v$ and
$c_1 \; \mathbf{op} \; v$.

\item If $\mathbf{op}$ is $\ne$ then
$\langle I,J \rangle$ is a solution of $\ell$ if and only if
either {\em (i)} $c > v$;  or 
{\em (ii)} $c < v$ and for every $p \in H_1$, 
$\pi(p) \ne v$.
\end{list}

\item{\sc Avg:}
Let $a = \frac{\sum_{p\in I} \{\pi(p)\}}{|I|}$
and $H_1 = {\cal H}(\ell)\setminus (I\cup J)$.

\begin{list}{$\bullet$}{\topsep=1pt \parsep=0pt \itemsep=1pt}
\item If $\mathbf{op}$ is $=$ then 
$\langle I,J\rangle$ is a solution of $\ell$ if and only if 
$a = v$ and for every $p \in H_1$, $\pi(p) = v$. This can be done
in time $O(|{\cal H}(\ell)|)$.

\item If $\mathbf{op} \in \{\geq,>\}$ then 
let $e_1,\dots, e_r$ be an enumeration of $H_1$ 
such that $\pi(e_i) \leq \pi(e_{i+1})$ for $1 \leq i\leq r-1$.
$\langle I,J\rangle$ is a solution of $\ell$ if and only if 
$a \;\mathbf{op}\; v$ and 
for each $ 0 \leq h \leq r$,
\[\sum_{p\in I}\pi(p) + \sum_{i=1}^h \pi(e_i) \;\mathbf{op}\; v\cdot |I|+ v \cdot h.\]
This can be accomplished in time $O(|{\cal H}(\ell)|^2)$.

\item If $\mathbf{op} \in \{\leq, <\}$ then
let $e_1,\dots, e_r$ be an enumeration of $H_1$ 
	such that $\pi(e_i) \geq \pi(e_{i+1})$ for $1 \leq i\leq r-1$.
$\langle I,J\rangle$ is a solution of $\ell$ if and only if 
$a \;\mathbf{op}\; v$ and 
 for each $ 0 \leq h \leq r$,
\[ \sum_{p\in I}\pi(p) + \sum_{i=1}^h \pi(e_i) \; \mathbf{op} \; v\cdot |I|+ v \cdot h. \]
This can be accomplished in time $O(|{\cal H}(\ell)|^2)$.
\end{list}
\end{itemize}
\end{proof}
The above lemma shows that there is a large class of programs 
with aggregates for which the problem of checking an answer set
and the problem of determining the existence of an answer set 
belongs to the class {\bf P} and {\bf NP} respectively. 

Observe that similar results can be extrapolated from the discussion
in Pelov's doctoral dissertation~\cite{Pelov04}.

\section{Conclusions and Future Work}

In this technical note, we defined $K^P_M$, a fixpoint operator for 
verifying answer sets of programs with aggregates. We showed that
the semantics for programs with aggregates described by this operator  
provides a new characterization of 
the semantics of \cite{SonPE05} 
for logic programs with aggregates. This operator converges to the
same semantics as in \cite{Pelov04} when ultimate approximating
aggregates are used.
We also related this semantics to recently proposed semantics
for aggregate programs. We discussed the complexity of the 
answer set checking problem and the problem of determining the existence
of an answer set. We showed that, for the class of programs with 
standard aggregates without the relation $\ne$ for {\sc Sum} and {\sc Avg}, 
the complexity of these two problems remains unchanged comparing 
to that of normal logic programs. In the future, we would like to 
use this idea in an efficient implementation of answer set solvers
with aggregates.

\subsubsection*{Acknowledgments}
The authors wish to thank the anonymous
reviewers for their insightful comments and for pointing out
relationships with existing literature, and Dr. Hing Leung for his
suggestions.

The research has been partially supported by NSF grants
HRD-0420407, CNS-0454066, and CNS-0220590.

\bibliographystyle{acmtrans}

\begin{thebibliography}{}

\bibitem[\protect\citeauthoryear{Cormen et al.}{Cormen et al.}{2001}]{cormen}
{\sc Cormen, T.H., Leiserson, C.E., Rivest, R.L.} {\sc and} {\sc Stein, C.} 2001.
\newblock {\em {Introduction to Algorithms, 2nd Edition}}.
\newblock MIT Press, Cambridge, MA.

\bibitem[\protect\citeauthoryear{Dovier, Pontelli, and Rossi}{Dovier
  et~al\mbox{.}}{2001}]{DovierPR01}
{\sc Dovier, A.}, {\sc Pontelli, E.}, {\sc and} {\sc Rossi, G.} 2001.
\newblock Constructive negation and constraint logic programming with sets.
\newblock {\em New Generation Comput.\/}~{\em 19,\/}~3, 209--256.

\bibitem[\protect\citeauthoryear{Dovier, Pontelli, and Rossi}{Dovier
  et~al\mbox{.}}{2003}]{DovierPR03}
{\sc Dovier, A.}, {\sc Pontelli, E.}, {\sc and} {\sc Rossi, G.} 2003.
\newblock Intensional Sets in CLP.
\newblock In {\em International Conference on Logic Programming},
	Springer, 284--299.

\bibitem[\protect\citeauthoryear{Elkabani, Pontelli, and Son}{Elkabani
  et~al\mbox{.}}{2004}]{ElkabaniPS04}
{\sc Elkabani, I.}, {\sc Pontelli, E.}, {\sc and} {\sc Son, T.~C.} 2004.
\newblock Smodels with CLP and its Applications: a Simple and Effective
  Approach to Aggregates in ASP.
\newblock In {\em International Conference on Logic Programming},
	Springer,  73--89.

\bibitem[\protect\citeauthoryear{Faber, Leone, and Pfeifer}{Faber
  et~al\mbox{.}}{2004}]{faberLP04}
{\sc Faber, W.}, {\sc Leone, N.}, {\sc and} {\sc Pfeifer, G.} 2004.
\newblock Recursive Aggregates in Disjunctive Logic Programs: Semantics and
  Complexity.
\newblock In {\em JELIA}, Springer, 200--212.


\bibitem[\protect\citeauthoryear{Gelfond}{Gelfond}{2002}]{a-prolog}
{\sc Gelfond, M.} 2002.
\newblock {Representing Knowledge in A-Prolog}.
\newblock In {\em Computational Logic: Logic Programming and Beyond},
  Springer Verlag, 413--451.

\bibitem[\protect\citeauthoryear{Gelfond and Lifschitz}{Gelfond and
  Lifschitz}{1988}]{GL88}
{\sc Gelfond, M.} {\sc and} {\sc Lifschitz, V.} 1988.
\newblock The Stable Model Semantics for Logic Programming.
\newblock In {\em International
  Conf.~and Symp. on Logic Programming}, MIT Press, 1070--1080.

\bibitem[\protect\citeauthoryear{Kemp and Stuckey}{Kemp and
  Stuckey}{1991}]{KempS91}
{\sc Kemp, D.~B.} {\sc and} {\sc Stuckey, P.~J.} 1991.
\newblock Semantics of Logic Programs with Aggregates.
\newblock In {\em ISLP}, MIT Press, 387--401.

\bibitem[\protect\citeauthoryear{Lloyd}{Lloyd}{1987}]{lloyd}
{\sc Lloyd, J.} 1987.
\newblock \emph{Foundations of Logic Programming}.
\newblock Springer Verlag.

\bibitem[\protect\citeauthoryear{Mumick, Pirahesh, and Ramakrishnan}{Mumick
  et~al\mbox{.}}{1990}]{MumickPR90}
{\sc Mumick, I.~S.}, {\sc Pirahesh, H.}, {\sc and} {\sc Ramakrishnan, R.} 1990.
\newblock The Magic of Duplicates and Aggregates.
\newblock In {\em Int. Conf. on Very Large Data Bases},
Morgan Kaufmann,
  264--277.

\bibitem[\protect\citeauthoryear{Pelov}{Pelov}{2004}]{Pelov04}
{\sc Pelov, N.} 2004.
\newblock {Semantic of Logic Programs with Aggregates}.
\newblock Ph.D. thesis, Katholieke Universiteit Leuven.


\bibitem[\protect\citeauthoryear{Pelov, Denecker, and Bruynooghe}{Pelov
  et~al\mbox{.}}{2003}]{PelovDB03}
{\sc Pelov, N.}, {\sc Denecker, M.}, {\sc and} {\sc Bruynooghe, M.} 2003.
\newblock {Translation of Aggregate Programs to Normal Logic Programs}.
\newblock In {\em {ASP: Advances in Theory and
  Implementation}}, CEUR Workshop  Proceedings. {29--42}.

\bibitem[\protect\citeauthoryear{Pelov, Denecker, and Bruynooghe}{Pelov
  et~al\mbox{.}}{2004}]{PelovDB04}
{\sc Pelov, N.}, {\sc Denecker, M.}, {\sc and} {\sc Bruynooghe, M.} 2004.
\newblock Partial Stable Models for Logic Programs with Aggregates.
\newblock In {\em LPNMR},
 Springer,
  207--219.

\bibitem[\protect\citeauthoryear{Ross and Sagiv}{Ross and
  Sagiv}{1997}]{RossS97}
{\sc Ross, K.~A.} {\sc and} {\sc Sagiv, Y.} 1997.
\newblock Monotonic Aggregation in Deductive Database.
\newblock {\em J. Comput. Syst. Sci.\/}~{\em 54,\/}~1, 79--97.

\bibitem[\protect\citeauthoryear{Son, Pontelli, and Elkabani}{Son
  et~al\mbox{.}}{2005}]{SonPE05}
{\sc Son, T.~C.}, {\sc Pontelli, E.}, {\sc and} {\sc Elkabani, I.} 2005.
\newblock {A Translational Semantics for Aggregates in Logic Programming }.
\newblock Tech. Rep. {CS-2005-006}, New Mexico State University.
\newblock \url{www.cs.nmsu.edu/CSWS/php/techReports.php?rpt_year=2005}.

\bibitem[\protect\citeauthoryear{Zaniolo, Arni, and Ong}{Zaniolo
  et~al\mbox{.}}{1993}]{ZanioloAO93}
{\sc Zaniolo, C.}, {\sc Arni, N.}, {\sc and} {\sc Ong, K.} 1993.
\newblock Negation and Aggregates in Recursive Rules: the LDL++ Approach.
\newblock In {\em DOOD}. 204--221.

\end{thebibliography}

\section*{Appendix A --- Correspondence between $K_M^P$ and $\Phi^{aggr}_P$}

We assume that 
the readers are familiar with the notations and definitions 
introduced in  \cite{PelovDB04}.

The three-valued immediate consequence operator 
$\Phi_P^{aggr}$ of a program $P$ in \cite{PelovDB04}, 
maps 3-valued interpretations to 3-valued interpretations. But 
3-valued interpretations can be split up in pairs $(I, J)$ 
of two valued interpretations such that $I \subseteq J$. 
Hence, an operator $\Phi_P^{aggr}$ can be viewed as an operator 
from pairs $(I,J)$ to pairs $\Phi_P^{aggr}(I,J) = (I',J')$ 
of 2-valued interpretations. It follows 
that $\Phi_P^{aggr}$ determines two component operators 
$\Phi_P^{aggr,1} (I, J) = I'$ and 
$\Phi_P^{aggr,2} (I, J) = J'$. The correspondence between 
$K_M^P$ and $\Phi^{aggr}_P$ is shown in the following claim.

\medskip \noindent 
{\bf Claim.} For every $I \subseteq M$, 
$K_M^P(I) = \Phi^{aggr,1}_P(I,M)$.

\begin{proof}
First, let us identify the aggregate atoms 
$agg(s) \;\; \mathbf{op} \;\; v$ 
in this paper with aggregate atoms $R(s, v)$ 
of \cite{PelovDB04}. E.g., {\sc Max}$(s) = v$ 
corresponds to {\sc Max}$(s, v)$; {\sc Max}$(s) \le v$ 
corresponds to {\sc Max}$_\le (s, v)$.  
Now we compare the definition of $K_M^P$ and $\Phi^{aggr,1}_P$
in the case that $I \subseteq M$. 
For simplicity let us assume that atom $a$ is defined by only one 
ground rule, say $r$.
 
$a \in K_M^P(I)$ iff $pos(r)$ is true in $I$, 
$neg(r)$ is false in $M$, and for each $\ell \in aggr(r)$, 
$l$ has a solution 
$(I \cap M \cap {\cal H}(\ell), {\cal H}(\ell) \setminus M)$. 

$a \in \Phi^{aggr,1}_P(I,M)$ iff $pos(r)$ is true in $I$, 
$neg(r)$ is false in $M$, and for each $\ell \in aggr(r)$,
$l$ evaluates to true, i.e., if 
$U^1_R(s^{(I,M)})) = t$. Here, $U^1_R$ is the first component of 
the three-valued aggregate, and $s^{(I,M)}$ is the 
evaluation of the set expression under the 3-valued interpretation 
$(I,M)$.

All that remains to be done is to show that 
$(I \cap M \cap {\cal H}(\ell), {\cal H}(\ell) \setminus M)$ is a 
solution for $l$ iff $U^1_R(s^{(I,M)}) = t$. 
Recall that we are considering the case where 
$I \subseteq M$, therefore the first expression simplifies to
$(I \cap {\cal H}(\ell), {\cal H}(\ell) \setminus M)$. 

Let us focus on set aggregates but the argument for multisets 
is the same. Let us consider an aggregate atom 
$$\ell = agg(s)\;\; \mathbf{op} \;\;v$$ 
where $$s = \{X \mid p(d_1,\ldots,d_{i-1},X, d_{i+1},\ldots,d_n)\}$$ 
and $X$ is the only variable and 
$d_1,\ldots,d_n$ are members of the Herbrand universe. 
For any $I \subseteq M$, \\
\begin{center}
$(I \cap {\cal H}(\ell), {\cal H}(\ell) \setminus M)$ 
is a solution for $\ell$ \\
iff for each $J$ such 
that 
$I \cap {\cal H}(\ell) \subseteq J$ and 
$J \cap ({\cal H}(\ell) \setminus M) = \emptyset$, 
$J \models \ell$ \\
iff for each $J$ such that $I \subseteq J \subseteq M$, 
$J \models \ell$. 
\end{center}
The latter equivalence is perhaps not entirely trivial 
but it follows easily from the fact that $J \models \ell \Leftrightarrow 
J'\models \ell$ whenever $J \cap {\cal H}(\ell) = J' \cap {\cal H}(\ell)$.
 
In \cite{PelovDB04}, the value $s^{(I,M)}$ 
is a three-valued (multi-)set, which can be written as a pair of 
two valued sets $(S_1, S_2)$ where 
$$S_1 = \{d \mid I \models p(d_1,\ldots,d_{i-1},d,d_{i+1},\ldots,d_n)\}$$ 
and 
$$S_2 = \{d \mid M \models p(d_1,\ldots,d_{i-1},d,d_{i+1},\ldots,d_n)\}.$$  

By definition of $U^1_R$, 
$U^1_R(s^{(I,M)})) = t$ iff for each set $S$ such that 
$S_1 \subseteq S \subseteq S_2$, $R(S, v)$ is true. 
It is straightforward to see that the conditions in this paragraph 
and the previous one are equivalent. 
\end{proof}

\end{document}